\documentclass[lettersize,journal]{IEEEtran}
\usepackage{amsmath,amsfonts}
\usepackage{algorithmic}
\usepackage{algorithm}
\usepackage{array}
\usepackage[caption=false,font=normalsize,labelfont=sf,textfont=sf]{subfig}
\usepackage{textcomp}
\usepackage{stfloats}
\usepackage{url}
\usepackage{verbatim}
\usepackage{graphicx}
\usepackage{cite}
\usepackage{multirow}
\usepackage{color}
\begin{document}

\title{PESFormer: Boosting Macro- and Micro-expression Spotting with Direct Timestamp Encoding}

\author{Wang-Wang~Yu,
        Kai-Fu Yang,
        Xiangrui Hu,
        Jingwen Jiang,
        Hong-Mei Yan,
        and~Yong-Jie Li,~\IEEEmembership{Member,~IEEE}
\thanks{W. Yu, K. Yang, X. Hu, H. Yan and Y. Li are with MOE Key Lab for Neuroinformation, 
        University Of Electronic Science And Technology Of China, Chengdu, China. 
        (Email: yuwangwang91@163.com, yangkf@uestc.edu.cn, huxiangrui2020@163.com, hmyan@uestc.edu.cn, liyj@uestc.edu.cn).
        }
\thanks{J. Jiang is with West China Biomedical Big Data Center, West China Hospital, 
        Sichuan University, Chengdu, China. 
        (Email: jiangjingwen@wchscu.cn).
        }
\thanks{Manuscript received September 30, 2024; revised August 16, 2021.(Corresponding author: Yong-Jie Li.)}}

\markboth{Journal of \LaTeX\ Class Files,~Vol.~14, No.~8, August~2021}%
{Shell \MakeLowercase{\textit{et al.}}: A Sample Article Using IEEEtran.cls for IEEE Journals}

\IEEEpubid{0000--0000/00\$00.00~\copyright~2021 IEEE}

\maketitle

\begin{abstract}
The task of macro- and micro-expression spotting aims to precisely localize and categorize temporal expression instances within untrimmed videos. Given the sparse 
distribution and varying durations of expressions, existing anchor-based methods often represent instances by encoding their deviations from predefined anchors. 
Additionally, these methods typically slice the untrimmed videos into fixed-length sliding windows.
However, anchor-based encoding often fails to capture all training intervals, and slicing the original video as sliding windows can result in valuable training intervals being discarded. 
To overcome these limitations, we introduce PESFormer, a simple yet effective model based on the vision transformer architecture to achieve point-to-interval expression spotting. 
PESFormer employs a direct timestamp 
encoding (DTE) approach to replace anchors, enabling binary classification of each timestamp instead of optimizing entire ground truths. Thus, all training intervals 
are retained in the form of discrete timestamps. To maximize the utilization of training intervals, 
we enhance the preprocessing process by replacing the short videos produced through the sliding window method.
Instead, we implement a strategy that involves zero-padding the untrimmed training videos to create uniform, longer videos of a predetermined duration. 
This operation efficiently preserves the original training intervals and eliminates video slice enhancement.
Extensive qualitative and quantitative evaluations on three datasets -- CAS(ME)$^2$, CAS(ME)$^3$ and SAMM-LV -- demonstrate that our PESFormer
outperforms existing techniques, achieving the best performance. 
\end{abstract}

\begin{IEEEkeywords}
Macro- and micro-expression spotting, vision transformer, direct timestamp encoding, fixed-duration with padding.
\end{IEEEkeywords}

\section{Introduction}
\IEEEPARstart{E}{xpression} is an important vehicle for conveying nonverbal information. It can be divided into two types: micro-expression (ME) and macro-expression (MaE) \cite{ekman1969nonverbal}.
The key characteristics of MEs, in contrast to MaEs, are short duration (lasting less than 0.5 second \cite{ekman2003darwin}), low intensity, and local motion \cite{wang2021mesnet}. 
These inherent complexities make it difficult to observe MEs, even for skilled professionals \cite{frank2009see}. Furthermore, MEs are more likely to manifest when individuals 
attempt to partially conceal or suppress their authentic emotions, whereas MaEs can sometimes convey misleading emotional cues. Given their significance, the analysis of both MEs and 
MaEs finds widespread application in high-stake scenarios, including medical diagnosis, criminal investigations, credit fraud detection, and public safety \cite{ekman2003darwin, ekman2009lie}. 

\begin{figure}[tbp]
  \centering
  \includegraphics[width=\linewidth]{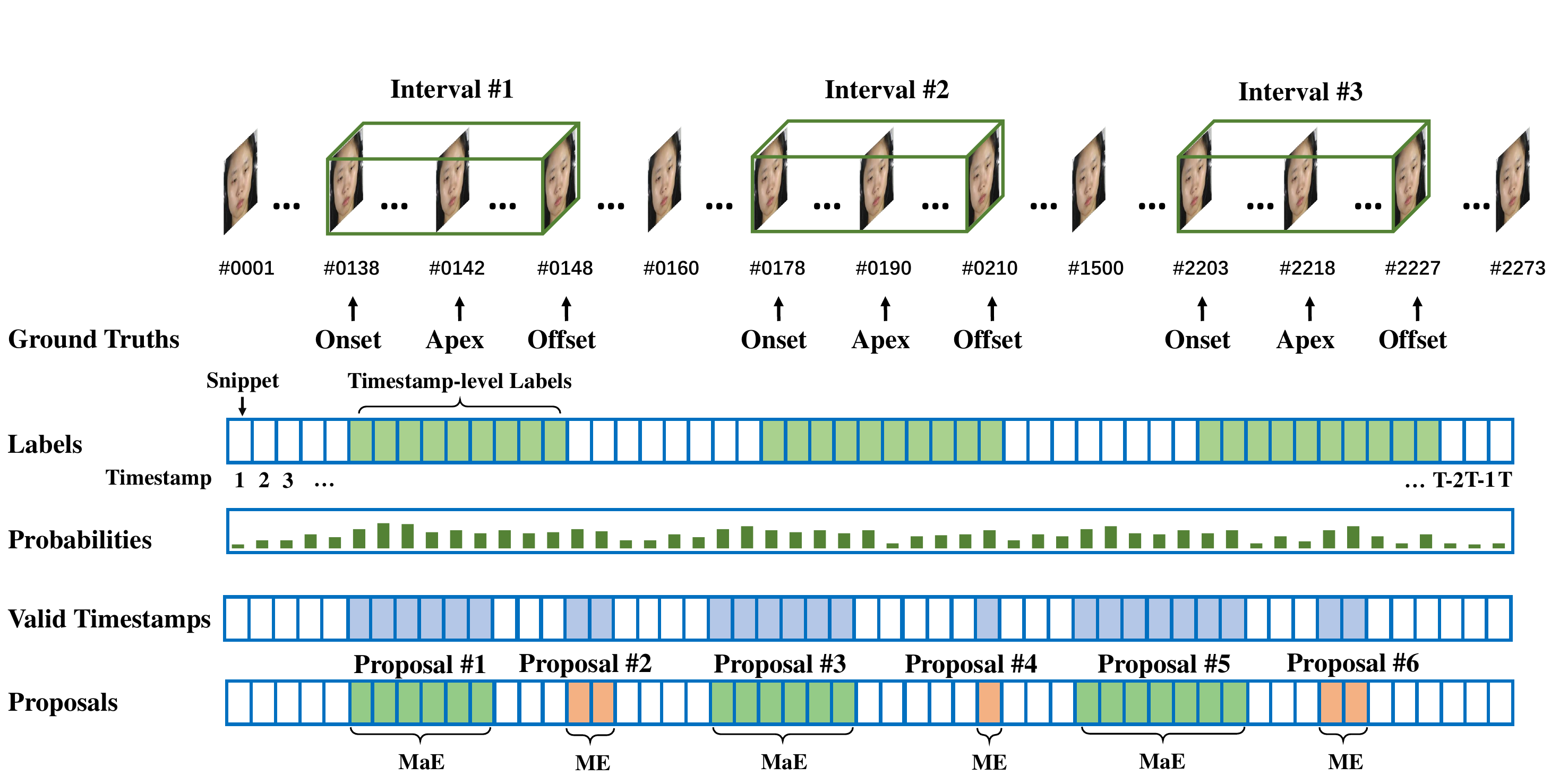}
  \caption{A long video from the CAS(ME)$^2$ dataset, spanning from frame \#1 to frame \#2273, undergoes a preprocessing stage where it is divided into a sequence of uniform timestamp snippets. 
  During training, each timestamp snippet contributes to the generation of a foreground probability, which is subsequently utilized in computing the loss. 
  During testing, these foreground probabilities serve as indicators to identify valid timestamp snippets. These valid timestamp snippets are then leveraged to generate proposals. 
  Our objective is to spot a set of consecutive video intervals that closely align with the ground truths.}
  \label{fig1}
\end{figure}  

Expression analysis comprises two key components: recognition and spotting. Recognition aims to categorize facial expressions into specific emotional groups \cite{xie2019adaptive} 
or assign them continuous multidimensional values \cite{du2014compound, li2017reliable, liang2020fine}. 
Expression spotting precedes recognition by determining whether a particular temporal instance corresponds to the background or contains a ME or MaE.
An expression can be delineated into three stages: onset, apex, and offset \cite{wang2021mesnet}. The onset marks the initial point, the apex represents the moment 
of maximum facial muscle deformation, reflecting the most prominent emotional information \cite{ekman1993facial, esposito2007amount}, and the offset signifies the ending. 
The existing spotting datasets including CAS(ME)$^2$\cite{qu2017cas}, SAMM-LV \cite{yap2020samm}, MMEW \cite{ben2021video}, and CAS(ME)$^3$ \cite{li2022cas} 
provide the onset, offset, and apex frames of all ground truths to facilitate model learning.

ME and MaE spotting has gained significant attention in the affective computing domain. Previous efforts have explored various approaches, including the utilization of anchor mechanisms 
to encode ground truths as deviations from predefined anchors within sliding windows \cite{yu2021lssnet, guo2023micro, yu2024Lgsnet}. Additionally, recurrent networks have been employed 
to capture temporal information from neighboring frames \cite{sun2019two, tran2019dense, verburg2019micro}. 
\IEEEpubidadjcol
Key frame-based methods have also been proposed to estimate the duration of 
expressions \cite{liong2021shallow, pan2020local, zhang2018smeconvnet, yap20213dcnn}. 
Furthermore, clip proposal networks have been developed to generate proposals for expression spotting \cite{wang2021mesnet}.
Despite the consistent advancements on key benchmarks \cite{qu2017cas, yap2020samm,li2022cas}, current methods often achieve accuracy at the expense 
of burdensome preprocessing steps and modeling complexities, including sliding window settings during preprocessing, anchor designs, network architectures, loss functions, output 
decoding during training, and proposal generation during testing.

Drawing from prior analysis, a clear pattern emerges: the essence of complexity is rooted in the effective encoding of comprehensive ground truths and the fine-grained generation 
of proposals. Achieving this requires a series of precise design and meticulous approaches to fine-tuning. To streamline the model's complexity, we propose a direct timestamp 
encoding (DTE) method to map ground truths to individual timestamps while concurrently identifying whether they belong to the foreground or background. In this method, 
as depicted in Figure \ref{fig1}, we define snippet as a group of continuous frames and timestamp as the serial number of each snippet.
By adopting this design, during training, we concentrate solely on classifying each timestamp, refraining from optimizing the onset and offset timestamps of each entire ground 
truth, practiced by existing methods \cite{yu2021lssnet, guo2023micro, yu2024Lgsnet}. 
Conversely, during testing, we meticulously combine only neighboring valid timestamps into proposals, ensuring both precision and efficiency.

In short, we develop a simple and effective transformer-based model, named PESFormer, for expression spotting. This approach is informed by the remarkable success of transformers 
in temporal action localization (TAL) \cite{zhang2022actionformer, xu2023boundary}. 
The main contributions of this work are summarized as follows: 
\begin{itemize}
  \item We rethink the necessity of intricate designs for expression spotting, and introduce a straightforward transformer-based model that comprises solely the fundamental components: 
  embedding block, transformer block, and output block. This simple approach enables us to only adopt a single classification output and two simple loss functions.
  \item Direct timestamp encoding (DTE) is designed to encode each timestamp without focusing on optimizing the deviations between ground truths and predefined anchors during training. 
  This significantly alleviates the challenges associated with model learning and proposal generation.
  \item During preprocessing, we set a larger fixed duration and complement with zero padding for the untrimmed videos as uniform length of training videos, instead of the 
  current way of splitting the original videos to generate the short-duration sliding windows. 
  This helps all relevant ground truths be retained within the training videos and minimize the loss of training intervals.
  \item The proposed method achieves state-of-the-art results on the CAS(ME)$^2$, CAS(ME)$^3$ and SAMM-LV datasets, demonstrating the effectiveness of our proposed approach in handling 
  the task of expression spotting.
\end{itemize}

\section{Related Work}

\subsection{Temporal Action Localization}
ME and MaE spotting in untrimmed videos involves classifying temporal intervals, similar to the task of temporal action localization (TAL) which is categorized as anchor-based or 
actionness-guided. Anchor-based methods encode ground truths as deviations from predefined anchors and refine them in sliding windows. For example, SSAD \cite{lin2017single} relies 
on anchors for classification and regression. TALNet \cite{chao2018rethinking} and R-C3D \cite{xu2017r} adopt Faster R-CNN \cite{ren2015faster}. A2Net \cite{yang2020revisiting} 
efficiently constructs a network with multiscale anchor-based and anchor-free branches, complementing each other. ActionFormer \cite{zhang2022actionformer} proposes a transformer-based 
model which uses a multi-level pyramid network and an anchor mechanism to achieve precise localization.

Actionness-guided TAL methods generate proposals through processing frame- and snippet-level confidence scores for the starting, ending, and duration. SSN \cite{zhao2017temporal} calculates 
actionness distributions and pools proposals to localize actions. BSN \cite{lin2018bsn} retrieves proposals using probabilities for starting, ending, and actionness. BMN \cite{lin2019bmn} 
predicts boundary probabilities and confidence maps to generate proposals. BSN$++$ \cite{su2021bsn++} further enriches context for boundary prediction with a boundary complementary 
classifier. TCANet \cite{qing2021temporal} incorporates temporal context to generate and refine proposals using regressors.

In general, anchor-based methods heavily rely on a significant number of predefined fixed anchors to achieve fine-grained localization \cite{lin2021learning}, but can't cover 
all ground truth intervals \cite{lin2021learning}. In contrast, actionness-guided models often capture more action instances but contain more false positive samples. To this end, we 
propose DTE in this work to encode each timestamp without emphasizing the precise onset and offset timestamps of individual ground truths, which can significantly alleviate the 
difficulties associated with model learning and proposal generation.

\begin{figure*}[tbp]
  \centering
  \includegraphics[width=\linewidth]{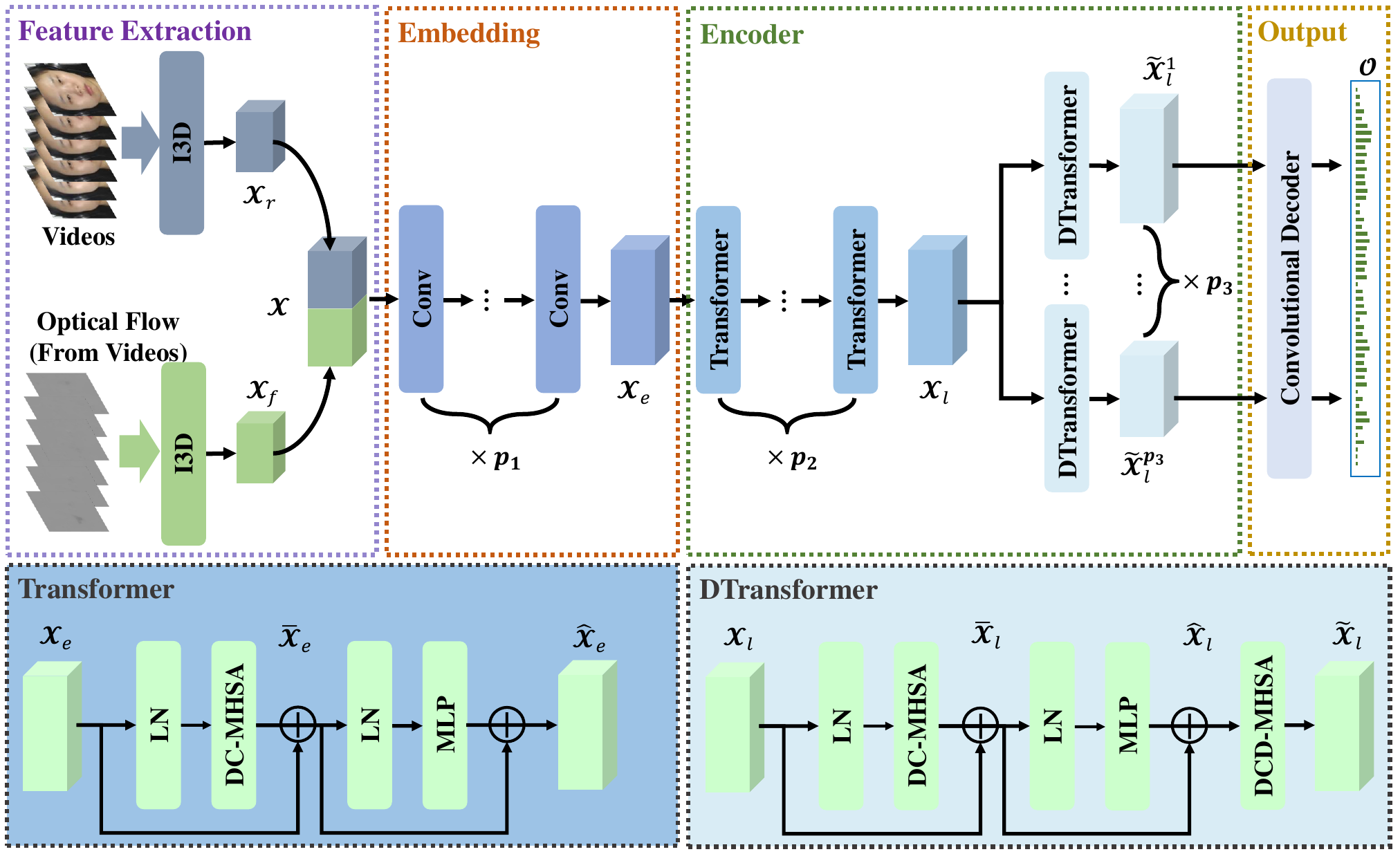}
  \caption{Overall schematic of PESFormer. 
  Given a video, we extract video features $\mathcal{X}_{r}$ and optical flow features $\mathcal{X}_{f}$ using a two-stream Inflated 3D ConvNets (I3D) model \cite{carreira2017quo}. 
  During training, these features are derived from a set of uniformly sampled, overlapping snippets from the video and its corresponding optical flow. The input features $\mathcal{X}$ 
  are formed by concatenating $\mathcal{X}_{r}$ and $\mathcal{X}_{f}$. Next, the input features $\mathcal{X}$ undergo $p_1$ convolution layers in the embedding component, resulting 
  in embedded features $\mathcal{X}_e$. These embedded features $\mathcal{X}_e$ are then processed by $p_2$ transformer networks in the temporal encoding component, generating fine-grained 
  features $\mathcal{X}_l$. To capture multiscale temporal information, we deploy $p_3$ downsampling transformer (DTransformer) networks, creating a feature pyramid network. The outputs 
  of this pyramid are $[\widetilde{\mathcal{X}}_l^1, ..., \widetilde{\mathcal{X}}_l^{p_3}]$, which are subsequently used to produce snippet-level probabilities $\mathcal{O}$ to indicate 
  the likelihood of each snippet belonging to the foreground. During testing, snippet-level probabilities $\mathcal{O}$ are employed to identify valid 
  snippets, which are subsequently combined to form expression proposals.
  }
  \label{fig2}
\end{figure*}  

\subsection{Vision Transformer}
Transformers, initially designed for natural language processing (NLP) \cite{vaswani2017attention, devlin2018bert}, have become successful in vision tasks \cite{dosovitskiy2020image, 
liu2021swim, liu2022swim, touvron2021training, wang2021pyramid}. Vision Transformer (ViT) \cite{dosovitskiy2020image} is the first pure transformer approach to set a new benchmark in 
image classification. SwimTransformer \cite{liu2021swim}, as a novel ViT model, achieves efficient local self-attention computation through hierarchical design and shift-window mechanism. 
Transformer models have also been explored in object detection \cite{carion2020end, dai2021dynamic}, segmentation \cite{wang2021end, cheng2021per}, and video 
understanding \cite{zhang2022actionformer, yang2022tubedetr}. As for the task of expression spotting, Guo et al. \cite{guo2023micro} make the first attempt to adopt transformer block 
\cite{dosovitskiy2020image} to replace one convolution layer to capture more fine-grained expression information. However, this simple substitution in deeper layers is not quite 
effective in improving the results of spotting. In contrast, we present a pure transformer-based model which can effectively improve the performance of spotting by well
capturing long-range dependency of expression in temporal dimension.

\subsection{Macro- and Micro-expression Spotting}
Based on the approaches used to generate proposals, deep learning-based expression spotting can be categorized into two main groups: key frame-based and interval-based methods.
Key frame-based techniques focus on localizing expression intervals by examining one or several frames within a long untrimmed video. For instance, Pan et al. \cite{pan2020local} 
categorize each frame into MaE, ME, or background. SMEConvNet \cite{zhang2018smeconvnet} uses a single frame to detect intervals. Yap et al. \cite{yap20213dcnn} rely on a fixed set
of durations, generating a significant number of negative samples. SOFTNet \cite{liong2021shallow}, on the other hand, introduces a shallow optical flow three-stream CNN model 
and pseudo-labeling to predict frame-level expressions. MC-WES \cite{yu2023weakly} and PWES \cite{yu2024weak} utilize weak labels during training, while selecting valid timestamps 
to generate proposals during testing. In general, these key frame-based spotting techniques have the disadvantage of missing numerous positive samples.

Interval-based methods, in contrast, consider all image features in the video and emphasize information from neighboring frames. Therefore, these techniques often utilize long 
short-term memory (LSTM) to encode temporal information \cite{sun2019two, tran2019dense, verburg2019micro}. However, LSTM has limitations in handling intricate and longer temporal 
patterns. Wang et al. \cite{wang2021mesnet} overcome this by employing a clip proposal network to establish long-range temporal dependencies. LSSNet \cite{yu2021lssnet} further 
improves upon this by incorporating both anchor-based and anchor-free branches. LGSNet \cite{yu2024Lgsnet} fuses local and global information to generate effective
proposals based on anchor mechanism.

In this paper, we propose the utilization of DTE as an innovative approach to encode each timestamp snippet, thereby facilitating binary classification without the necessity of 
optimizing the deviations between ground truths and predefined anchors during training. 
This approach streamlines the process and enhances efficiency in handling temporal expression spotting.

\section{Method}

\subsection{Problem Formulation}
\label{Section 3.1}
Consider a training video $\mathcal{S} =\{{S_l}\}_{l=1}^L$, which contains $L$ frames and $M$ expressions. The ground truth instances of these expressions are denoted by 
$\Psi = \{f_{on}^{(m)}, f_{off}^{(m)}, y^{(m)}\}_{m=1}^{M}$, where $f_{on}^{(m)}$ and $f_{off}^{(m)}$ represent the onset and offset frames, respectively, for the $m-$th expression 
instance belonging to the class $y^{(m)}$. The expression classes are represented as $y^{(m)}\in \mathbb{R}^{C}$, where $C$ is the total number of expression categories and 
$C=2$ in this work, indicating the two classes of ME and MaE.

The objective of the spotting task is to accurately identify and localize all potential proposals (expression intervals) within the untrimmed video. To facilitate this, during preprocessing, 
as shown in Figure \ref{fig1}, each training video is divided into a sequence of snippets of equal length marked with timestamps, which are used for the entire training and 
testing of the model. During testing, the model generates predicted proposals based on these snippets with timestamps, which are then decoded into frame-level results. Each proposal $i$ is 
represented as ${E}^{(i)} = (f_{on}^{(i)}, f_{off}^{(i)}, y^{(i)}, \phi^{(i)})$. Here, $f_{on}^{(i)}$ indicates the onset frame, $f_{off}^{(i)}$ denotes the offset frame, $y^{(i)}$ specifies 
the categorical label, and $\phi^{(i)}$ quantifies the confidence score.

\subsection{Model Structure}
\label{Section 3.2}
As shown in Figure \ref{fig2}, the proposed model, named PESFormer, is built upon the ViT architecture. A two-stream Inflated 3D ConvNets (I3D) model \cite{carreira2017quo} is first employed 
to extract features from both the original videos and the deduced optical flow. Subsequently, $p_1$ embedding layers are utilized to further refine these extracted features. To enhance the 
identification of proposals across multiple temporal scales, a temporal encoder that integrates $p_2$ transformer networks is specifically designed to optimize features originating from diverse 
layers. Finally, a shared decoder, equipped with $p_3$ downsampling transformer (DTransformer) networks, generates the prediction outcomes for both classification and localization tasks.

\subsubsection{Feature Extraction}
\label{Section 3.2.1}
\begin{figure}[tbp]
  \centering
  \includegraphics[width=\linewidth]{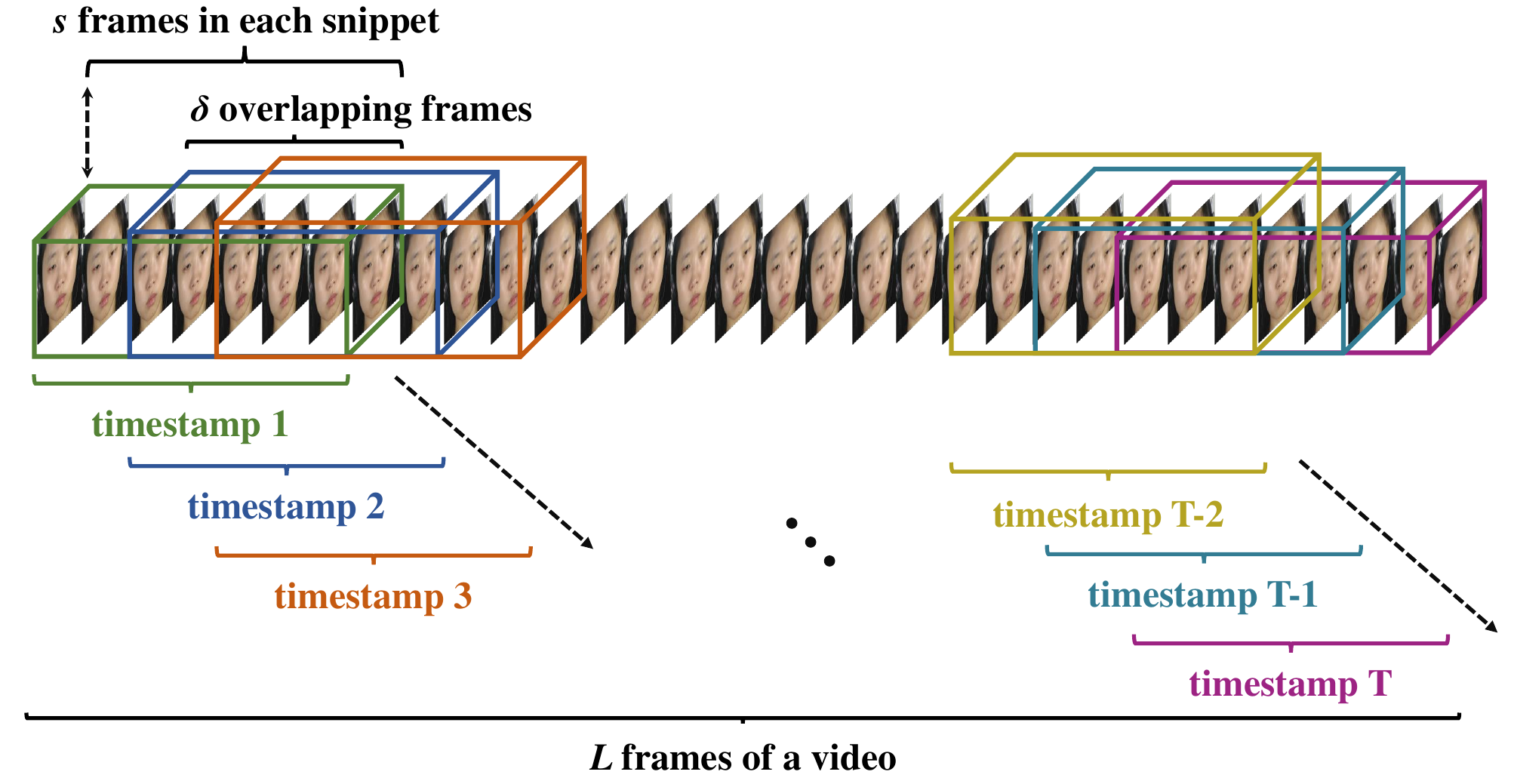}
  \caption{Snippet extraction process. 
  Given a video that comprises a total of $L$ frames, we split it into timestamp snippet, each consisting of consecutive $s$ frames. The overlap between neighboring timestamp
  snippets are $\delta$ frames.
  }
  \label{fig3}
\end{figure}

Drawing inspiration from prior models \cite{yu2021lssnet, yu2024Lgsnet, yu2023weakly, yu2024weak}, we first generate optical flow from the videos using the TVL1 algorithm \cite{wedel2009improved} 
with the default smoothing parameter ($\lambda=0.15$), which can obtain dense optical flow between adjacent frames. As shown in Figure \ref{fig3}, we divide both the videos and the optical flow 
into a series of overlapping uniform timestamp snippets, with adjacent snippets sharing $\delta$ overlapping frames, where each snippet comprises $s$ frames. To capture rich representations, we 
employ the I3D model to extract image features $\mathcal{X}_{r} \in \mathbb{R}^{T\times D}$ and optical flow features $\mathcal{X}_{f} \in \mathbb{R}^{T\times D}$ from these snippets. Here, 
$T=1+(L-s)/(s-\delta)$ represents the total number of snippets, and $D$ denotes the dimensionality of the features extracted from each snippet. Then, we concatenate $\mathcal{X}_{r}$ and 
$\mathcal{X}_{f}$ as input features $\mathcal{X} \in \mathbb{R}^{T\times 2D}$. To maintain consistency between the number of video images and optical flow frames, we discard the last frame of the videos. 

\subsubsection{Feature Embedding}
\label{Section 3.2.2}
Given that each video is segmented into distinct timestamp snippets, to improve the adaptation of image patches in Vision Transformer (ViT) \cite{dosovitskiy2020image}, each snippet can 
be regarded as a video patch for patching. To maintain consistency with ViT, positional embedding is incorporated. However, existing research \cite{zhang2022actionformer} indicates that 
positional embedding can actually diminish model performance. Additionally, recent findings \cite{zhang2022actionformer, xiao2021early, li2019enhancing} have demonstrated the benefits 
of incorporating convolutions prior to the Transformer network. This approach effectively integrates local context from temporal data, thereby stabilizing the training of the visual 
transformer. Consequently, we configure $p_1$ convolutional layers with ReLU as embedding blocks, ultimately enhancing the granularity of input features $\mathcal{X}$. We denote the output
of this component as the embedded features $\mathcal{X}_e \in \mathbb{R}^{T\times D_e}$.

\subsubsection{Timestamp Encoder}
\label{Section 3.2.3}
As depicted in Figure \ref{fig2}, the timestamp snippet encoder comprises multiple transformer networks and downsampling transformer (DTransformer) networks, serving to refine the embedded features
$\mathcal{X}_e$ into $\mathcal{X}_l\in \mathbb{R}^ {T \times D_e}$ and extract multiscale temporal information from $\mathcal{X}_l$, respectively. Within this component, the transformer network incorporates 
multi-head self-attention (MHSA) blocks and multilayer perceptron (MLP) blocks, each preceded by layer normalization and followed by residual connections, adhering to the practices 
established in \cite{wang2019learning, baevski2019adaptive}. As for the DTransformer, an additional downsampling block is incorporated compared with the former to facilitate the creation 
of multiscale temporal features.

\noindent \textbf{Self-attention Mechanism.} 
Among the above blocks, MHSA leverages the self-attention mechanism \cite{vaswani2017attention}. By computing the correlation scores between each element within a sequence and the 
remaining elements, long-range dependencies can be captured, which will facilitate efficient comprehension and representation of the input sequence. MHSA is an 
extension of self-attention mechanisms in which multiple self-attention operations are run in parallel. 

In detail, the embedded features $\mathcal{X}_e$ is processed with three convolution layers $\mathcal{W}_q \in \mathbb{R}^{D_e \times D_e}$, 
$\mathcal{W}_k \in \mathbb{R}^{D_e \times D_e}$, $\mathcal{W}_v \in \mathbb{R}^{D_e \times D_e}$,
\begin{equation}
  \mathcal{Q} =\mathcal{X}_e \mathcal{W}_q ,\mathcal{K} = \mathcal{X}_e \mathcal{W}_k ,\mathcal{V} = \mathcal{X}_e \mathcal{W}_v,
  \label{equ3.1}
\end{equation}
where $\mathcal{Q} \in \mathbb{R}^{T \times D_e}$, $\mathcal{K} \in \mathbb{R}^{T \times D_e}$, $\mathcal{V} \in \mathbb{R}^{T \times D_e}$ are feature representations, referred to as 
query, key and value respectively in the self-attention mechanism \cite{vaswani2017attention}. The query $\mathcal{Q}$ and key $\mathcal{K} $ are used to calculate the attention weights 
$\mathcal{A} \in \mathbb{R}^{T \times T}$, which is then multiplied with value $\mathcal{V}$,
\begin{equation}
  \mathcal{A} = softmax(\mathcal{Q} \mathcal{K}^\top /\sqrt{D_e}), 
  \label{equ3.2}
\end{equation}
\begin{equation}
  \mathcal{SA}(\mathcal{X}_e) = \mathcal{A} \mathcal{V},
  \label{equ3.3}
\end{equation}
where $\mathcal{SA}$ is the result of self-attention operations. 

\noindent \textbf{Multi-head Self-attention (MHSA) Mechanism.} 
To optimize the execution of the aforementioned parallel operations, we adopt the innovative approach introduced by ActionFormer \cite{zhang2022actionformer}, 
integrating $N$ attention heads within the MHSA block. This configuration allows 
the MHSA block to efficiently handle multiple attention mechanisms simultaneously, leading to improved overall performance. In specific, the MHSA block is simply defined as,
\begin{equation}
  \mathcal{MHSA}(\mathcal{X}_e) = [\mathcal{SA}_1(\mathcal{X}_{e}^1),..., \mathcal{SA}_n(\mathcal{X}_{e}^n),...,\mathcal{SA}_N(\mathcal{X}_{e}^N)],
  \label{equ3.4}
\end{equation}
where $\mathcal{X}_{e}^n\in \mathbb{R}^{T\times D_e/N}$ is the $n$-th subspace of the embedded features and $\mathcal{MHSA}(\mathcal{X}_e)$ is the final representation. MHSA's key concept 
involves dividing a linear transformation into multiple heads. Each head independently performs self-attention operation, computing the subspace attention weights $\mathcal{A}_n\in\mathbb{R}^{T \times T}$ 
and corresponding outputs $\mathcal{SA}_n(\mathcal{X}_{e}^n)$. Subsequently, these subspace outputs are merged to create the final representation $\mathcal{MHSA}(\mathcal{X}_e)$. This approach 
allows the model to capture crucial information across diverse subspaces, thereby enhancing its comprehension of complex data. Furthermore, MHSA's self-attention operation effectively identifies 
dependencies within input sequences, leading to a deeper understanding of contextual information and improved model performance. 

\noindent \textbf{Duration-constrained MHSA.} 
However, recent works \cite{zhang2022actionformer, choromanski2021rethinking} highlight that MHSA can significantly increase computational complexity and cost. To mitigate this, the concept of 
local temporal self-attention has emerged \cite{choromanski2021rethinking}, confining the attention mechanism to a local temporal window. 

Given that the duration of facial expression typically does not exceed 4.0 seconds \cite{ekman2003darwin} and that temporal contexts shorter than a certain local temporal range become less 
informative for capturing long-range dependencies \cite{zhang2022actionformer}, we introduce a local fixed length $H$ and then generate $U$ larger temporal subspaces by slicing the above $n$-th 
subspace along the temporal dimension into our MHSA blocks. The number $U$ of these larger temporal subspaces is calculated as $U =T/H$.

This approach allows us to capture longer local temporal dependencies while considering both upper and lower duration bounds, without relying on global temporal dependencies. The modified blocks, 
termed duration-constrained MHSA (DC-MHSA) blocks, serve to efficiently capture relevant temporal dependencies within the specified duration constraints. 
Therefore, the DC-MHSA block is defined as,
\begin{equation}
  \begin{split}
    \mathcal{DCMHSA}(\mathcal{X}_e) & = [\mathcal{SA}_{11}(\mathcal{X}_e^{11}), ..., \mathcal{SA}_{n1}(\mathcal{X}_e^{n1}),..., \\
    & \mathcal{SA}_{nu}(\mathcal{X}_e^{nu}),...,\mathcal{SA}_{NU}(\mathcal{X}_e^{NU})],
  \end{split}
  \label{equ3.5}
\end{equation}
where $\mathcal{X}_e^{nu}\in \mathbb{R}^{H \times D_e/N}$ represents the $nu$-th subspace of the embedded features, and $\mathcal{DCMHSA}(\mathcal{X}_e)$ denotes the final output of our DC-MHSA 
block. Through this design, our approach is able to capture richer local information and effectively integrate it with global temporal information. 

\noindent \textbf{Transformer.} 
The output of the transformer network can be expressed as,
\begin{equation}
  \overline{\mathcal{X}}_e^{nu} = \mathcal{DCMHSA}(\mathcal{LN}(\mathcal{X}_e^{nu})) + \mathcal{X}_e^{nu},
  \label{equ3.6}
\end{equation}
\begin{equation}
  \widehat{\mathcal{X}}_e^{nu} = \mathcal{MLP}(\mathcal{LN}(\overline{\mathcal{X}}_e^{nu})) + \overline{\mathcal{X}}_e^{nu} ,
  \label{equ3.7}
\end{equation}
where $\widehat{\mathcal{X}}_e \in \mathbb{R}^{T \times D_e}$ represents the output generated by the transformer network, $\mathcal{LN}$ denotes the layer-norm operation, which helps stabilize 
the training process by normalizing the layer outputs. $\mathcal{MLP}$ refers to the multilayer perceptron (MLP) network. This transformer network specifically comprises two convolution layers, 
interspersed with a GELU non-linearity. This combination of convolutions and GELU activation enhances the network's ability to learn complex representations. To further extract fine-grained 
features, we employ $p_2$ transformer networks.

\noindent \textbf{DTransformer.} 
Similar to LSSNet \cite{yu2021lssnet} and LGSNet \cite{yu2024Lgsnet}, we construct a feature pyramid network to effectively spot multiscale expressions. Consequently, we integrate downsampling blocks 
into the DTransformer networks to form duration-constrained downsampling MHSA (DCD-MHSA) blocks. These blocks enable us to appropriately reduce the temporal dimension while preserving the 
capacity to capture relevant long-range dependencies within specified duration constraints. 
Assuming that the output of the above transformer networks is $\mathcal{X}_l$,
\begin{equation}
  \widehat{\mathcal{X}}_l^{nu} = \mathbb{TN}(\mathcal{X}_l^{nu}),
  \label{equ3.8}
\end{equation}
where $\mathbb{TN}$ denotes the same transformation procedure as employed by the aforementioned transformer network, and the $\widehat{\mathcal{X}}_l^{nu}$ represents the output of $\mathbb{TN}$.
Thus, the DCD-MHSA block is defined as,
\begin{equation}
  \widetilde{\mathcal{X}}_l^{nu} = \mathcal{DCMHSA}(\widehat{\mathcal{X}}_l^{nu} \mathcal{W}_l),
  \label{equ3.9}
\end{equation}
where $\mathcal{W}_l \in \mathbb{R}^ {D_e \times D_e}$ is a downsampling convolution layer used to halve the temporal dimension of $\widehat{\mathcal{X}}_l$, and $\widetilde{\mathcal{X}}_{l}$ 
indicates the output of the DTransformer networks. As shown in Figure \ref{fig2}, to capture multiscale temporal information, we deploy $p_3$ DTransformer networks, creating 
a feature pyramid network. This innovative architecture allows us to extract and process information across diverse temporal scales, significantly enhancing the overall representational capacity of 
our model.

\subsubsection{Output Decoder}
These pyramid features $[\widetilde{\mathcal{X}}_l^1, ..., \widetilde{\mathcal{X}}_l^{p_3}]$ obtained from the DCD-MHSA blocks undergo further processing through a convolution layer 
and the sigmoid function, ultimately yielding snippet-level probabilities $\mathcal{O} \in \mathbb{R}^T$. These probabilities serve as indicators of the likelihood of each snippet belonging to the 
foreground. During testing, these snippet-level probabilities $\mathcal{O}$ are employed to identify valid snippets, which are subsequently combined to form the predicted expression proposal intervals.

\subsection{Direct Timestamp Encoding (DTE)}
\label{Section 3.3}
The core of an effective encoding approach lies in the accessibility of encoding comprehensive ground truths and the simplicity of generating fine-grained proposals. Currently, indirect encoding methods 
based on anchors are considered optimal \cite{yu2024Lgsnet}. The anchor mechanism, initially introduced in object detection, simplifies the process by utilizing predefined boxes tailored to the size and 
shape of potential objects \cite{zhong2020anchor, ren2015faster}. 
Subsequently, this concept has been expanded to temporal action localization (TAL) \cite{yang2020revisiting, chao2018rethinking, cheng2022tallformer, zhang2022actionformer}. It optimizes training 
data by ensuring a balanced representation of foreground and background examples. 
By adapting to varying foreground scales, the anchor mechanism enhances the model's flexibility and precision \cite{yang2018metaanchor}. 

\begin{figure}[tbp]
  \centering
  \includegraphics[width=\linewidth]{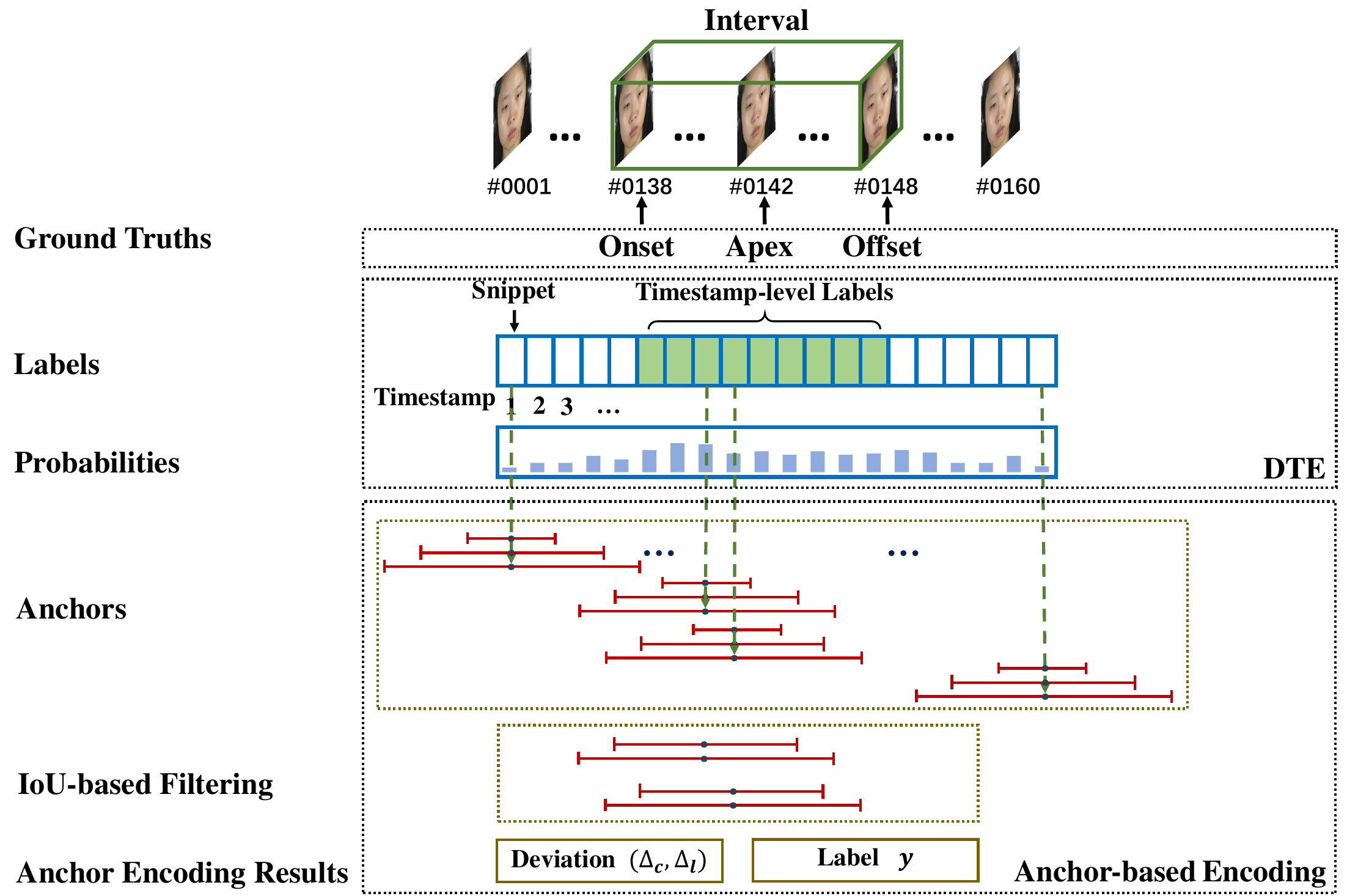}
  \caption{Illustration of the two encoding methods of the proposed DTE and the anchor-based encoding. The DTE method, meticulously assesses whether each individual timestamp snippet suffices to 
  qualify as part of the foreground, embodying a localized approach. Conversely, the anchor-based encoding identifies valid anchors by computing the intersection over union (IoU) between pre-defined 
  anchors and training intervals, subsequently selecting those anchors whose IoU exceeds a specified threshold. These valid anchors are utilized to encode training intervals in terms of center deviation 
  $\Delta_c$ and duration deviations $\Delta_l$. Each anchor's corresponding set of deviations from the training intervals is then assigned a class $y$ label derived from the training intervals themselves.}
  \label{fig4}
\end{figure}  

By incorporating anchor mechanism into expression spotting, existing methods \cite{yu2024Lgsnet, yu2021lssnet} encode each ground truth as deviations in location and duration relative to predefined 
anchors as depicted in Figure \ref{fig4}. Nevertheless, anchor-based methods rely on large fixed anchors to encode the majority of ground truths, which undoubtedly increases the difficulty of 
parameterization and steepens the learning curve. Furthermore, these methods commonly employ non-maximum suppression (NMS) to generate proposals during testing, making them computationally costly. 

Given the challenges posed by current anchor-based methods, we propose an efficient encoding approach that aims to simplify the learning process, eliminate tedious parameter tuning, and encode all 
ground truths in a straightforward manner. As illustrated in Figure \ref{fig4}, the proposed direct timestamp encoding (DTE) method treats each timestamp snippet as the smallest fine-grained unit within the 
temporal sequences. Unlike the anchor-based methods, our DTE method does not rely on predefined anchors to encode each ground truth as deviation groups with temporal location and durations and uses the 
deviations as the optimization objective. Instead, our approach simply maps each ground truth to the corresponding video sequence, assigning the timestamp snippet belonging to any ground truth as the foreground.
Consequently, there is also no need to filter encoding deviation groups by calculating the intersection over union (IoU). 

From this perspective, we also avoid the need to assign categories to sets of valid encoding deviation groups, as done in existing methods like LSSNet and LGSNet. Instead, our training process 
solely focuses on foreground-background classification for each timestamp snippet, eliminating the need to optimize onset and offset timestamps or compute category loss, as required by the anchor 
based approaches. During testing, we summarize only those timestamp snippets which are larger than a predefined value as valid ones into the proposal. Furthermore, by analyzing the 
duration of each proposal, we can accurately determine whether it belongs to MEs or MaEs \cite{ekman2003darwin}, ensuring both precision and efficiency. As a result, there is no requirement to 
filter proposals by utilizing NMS.

In this scenario, our DTE significantly alleviates the difficulties associated with model learning and proposal generation, making it a more robust and efficient encoding solution for 
expression spotting tasks.

\subsection{Model Training}
\label{Section 3.4}
As shown in Figure \ref{fig2}, the output of our PESFormer is a one-dimensional vector $\mathcal{O} \in \mathbb{R}^T$ that specifies the probability that each timestamp snippet belongs 
to the foreground. Since expressions often maintain in just a few timestamp snippets within a video, the distribution of foreground timestamp snippets tends to be sparse in 
most given videos \cite{xu2016heterogeneous}. To address the imbalance between foreground and background timestamps and refine the model's optimization, we incorporate the focal loss 
function \cite{lin2017focal} and the dice loss \cite{milletari2016v} function into our approach.

\noindent \textbf{Focal Loss.}
The focal loss function was specifically designed to tackle the problem of class imbalance, prevalent in object detection tasks where background samples significantly outweigh foreground 
samples \cite{lin2017focal}. By reducing the focus on easily classified samples and enhancing the learning of challenging ones, this loss function bolsters the model's ability to recognize 
minority classes, enabling it to handle imbalanced datasets effectively.

\noindent \textbf{Dice Loss.}
The dice loss function, alternatively known as the dice coefficient loss, is a widely used loss function in computer vision, particularly in medical image segmentation \cite{milletari2016v}. 
It also addresses the class imbalance challenge, a common issue in image segmentation where the pixel distribution between foreground and background can vary significantly. By optimizing the 
dice coefficients, this loss function sharpens the model's segmentation performance for small objects or minority classes.

In this paper, we incorporate both the focal loss function $\mathcal{L}_{fl}$ and the dice loss function $\mathcal{L}_{dl}$ to mitigate the imbalance between foreground and background 
timestamp snippets. 
\begin{equation}
  \mathcal{L}= \mathcal{L}_{fl} + \lambda \mathcal{L}_{dl},
  \label{equ3.10}
\end{equation}
where $\lambda$ is a predefined hyperparameter. This approach better highlights the differences between these snippets, enabling our model to discriminate more effectively. 

\subsection{Preprocessing with Zero-padding Based Large Fixed Duration}
\label{Section 3.5}
\begin{figure}[tbp]
  \centering
  \includegraphics[width=\linewidth]{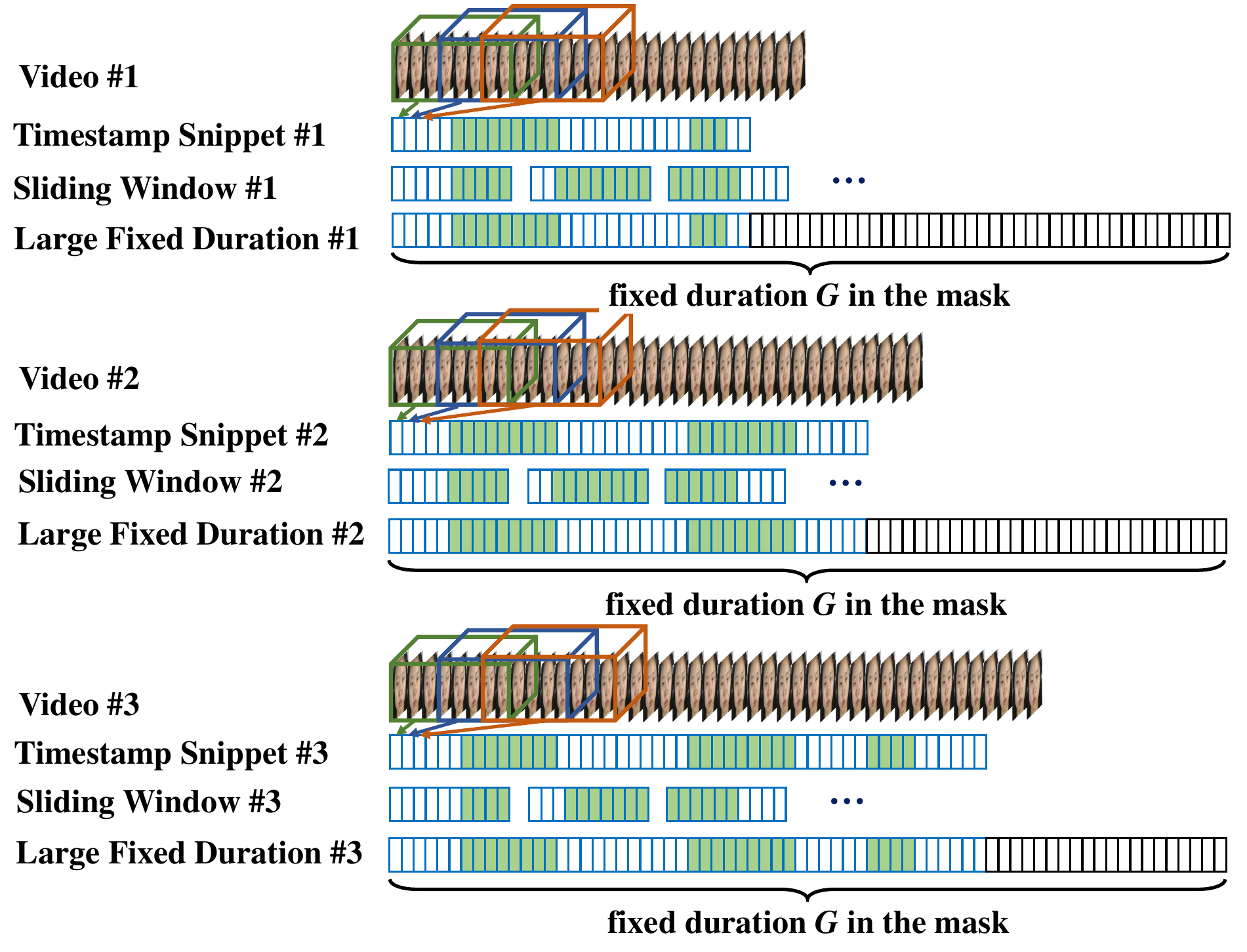}
  \caption{Three examples illustrating two different preprocessing methods for dividing long videos into multiple snippets. The first method is the sliding window approach, which involves dividing 
  untrimmed videos of varying durations into a substantial collection of shorter videos, with uniform length, specifically tailored for training purposes. Alternatively, our large fixed duration 
  method ensures duration consistency by appending zeros (zero-padding) to untrimmed videos of different durations, thereby 
  unifying their durations prior to utilization in the training process.}
  \label{fig5}
\end{figure}

As shown in Figure \ref{fig5}, sliding windows are widely employed in video understanding tasks \cite{yang2020revisiting, lee2021learning, paul2018w} to preprocess untrimmed videos by cropping 
them into uniform video clips. Such approach can ensure that at least one relatively complete ground truth action is retained, significantly reducing the search space. This reduction is beneficial 
for model convergence and proposal generation \cite{yu2021lssnet, yu2024Lgsnet}. However, such approach has its limitations. Firstly, given the brevity of MEs, if a sliding window fails to 
retain an ME during the slicing process, this sample would be lost. Secondly, while adopting an overlapping approach for slicing the original video may generate more training sliding windows, which 
inevitably increases computational cost. 

To address the above limitations of sliding windows in video preprocessing, we introduce a straightforward and efficient method similar to that of ActionFormer \cite{zhang2022actionformer}. As shown 
in Figure \ref{fig5}, our approach involves using a larger fixed duration $G$ for each training video and complementing the untrimmed videos with zero-padding. This ensures that all relevant ground 
truths are retained within the training videos, minimizing the loss of training intervals and ensuring consistent input dimension. To mitigate the potential interference caused by the padded regions, 
we further introduce a mask mechanism. This mask, composed of a sequence of $0$ and $1$, serves to indicate the original video content and the padded areas. This mask is utilized both during training 
and testing, effectively reducing the impact of zero-padding on the model's learning process.

\subsection{Proposal Generation}
\label{Section 3.6}
During testing, we use a truncation threshold $\theta$ to identify and select timestamp snippets with lengths larger than $\theta$ as valid timestamp snippets. Subsequently, we 
integrate these consecutive video 
timestamp snippets into potential proposals and subsequently map these proposals onto the original video, transforming them into frame-level proposals, labeled as $(f_{on}^{(i)},f_{off}^{(i)})$, where 
$f_{on}^{(i)}$ represents the onset frame and $f_{off}^{(i)}$ denotes the offset frame for the $i$-th proposal. Based on the duration spanned between $f_{on}^{(i)}$ and $f_{off}^{(i)}$, we categorize 
each proposal into $y^{(i)}$--those proposals lasting no more than 0.5 second are classified as MEs, while the rest are MaEs \cite{paul2007emotions}. Regarding the confidence score $\phi^{(i)}$, we adopt 
the same approach utilized in MC-WES\cite{yu2023weakly} and PWES\cite{yu2024weak} to calculate $\phi^{(i)}$ for the $i$-th proposal.

\section{Experiments}

\subsection{Datasets}
\label{Section 4.1}
We evaluate our proposed model using three widely recognized datasets for expression spotting: CAS(ME)$^2$ \cite{qu2017cas}, CAS(ME)$^3$ \cite{li2022cas}, and SAMM-LV \cite{yap2020samm}. 
CAS(ME)$^2$ comprises 98 videos, averaging 2940 frames per video captured at 30 FPS. Notably, 96\% of these frames are background. This dataset is meticulously annotated with 57 ME 
and 300 MaE intervals. CAS(ME)$^3$ stands out with its extensive collection of 956 videos, averaging 2600 frames per video at 30 FPS. Similarly, 84\% of its frames are background. This dataset 
is comprehensively annotated, featuring 207 MEs and 2071 MaEs. Lastly, SAMM-LV offers a collection of 224 longer videos, averaging 7000 frames per video captured at a higher rate of 200 FPS. 
This dataset contains 159 MEs and 340 MaEs, and the remaining approximate 68\% frames are background.

\subsection{Evaluation Metric}
\label{Section 4.2}
Following the Micro-Expression Grand Challenge (MEGC) in both 2019 \cite{see2019megc} and 2022 \cite{li2022megc2022}, a predicted expression proposal $E$ is deemed to be a true positive (TP) sample 
solely if it aligns with a ground truth interval $E_{gt}$ that fulfills the criteria outlined,
\begin{equation} 
  \frac{E\bigcap E_{gt}}{E\bigcup  E_{gt}} \geq k_{eval},
  \label{equ4.1}
\end{equation} 
where $k_{eval}$ is the evaluation threshold and set to 0.5.

\subsection{Implementation Details}
\label{Section 4.3}
As shown in Figure \ref{fig3}, on the CAS(ME)$^2$ and CAS(ME)$^3$ datasets, we sample continuous 8 frames ($s=8$) with $\delta = 6$ overlapping frames as timestamp snippets to divide the videos 
and generated optical flow. On the SAMM-LV dataset, we set $s=32$ with $\delta = 24$ overlapping frames, considering the quite higher FPS of this dataset. Subsequently, we leverage an I3D model 
to extract 1024-dimensional features from each snippet.

During training, we establish fixed duration $G$ for each dataset: CAS(ME)$^2$ is set to 2304 timestamp snippets, CAS(ME)$^3$ to 2592 snippets, and SAMM-LV to 2880 snippets. In cases where the 
duration of a training video fall short of these preset fixed durations, we pad the remaining snippets with zeros. For the CAS(ME)$^2$ and SAMM-LV datasets, 
the batch size is set to 4, while for CAS(ME)$^3$, the batch size is set to 16. We adopt Adam \cite{kingma2014adam} as optimizer and train for 30 epochs for each dataset. Additionally, both DC-MHSA 
and DCD-MHSA empirically have $N=4$ heads, enabling efficient MHSA blocks.

Regarding hyperparameters, we set $p_1 = p_2 = 2$, $p_3 = 1$, and the learning rate is fixed at 0.00001 for all three datasets. During testing, we adjust the truncation threshold $\theta$ 
as 0.05 on the CAS(ME)$^2$, CAS(ME)$^3$ and SAMM-LV datasets. We set $\lambda = 1.0$ in Equation \ref{equ3.10}, which will be analyzed in the following ablation study.

\subsection{Ablation Study}
We employ a leave-one-subject-out (LOSO) learning strategy for training our model and generating proposals. After training, we evaluate the model's performance by determining the recall rate, 
precision rate, and F1-score based on the retained proposals within three datasets. In accordance with the evaluation protocol outlined in LSGNet \cite{yu2024Lgsnet}, for the analysis of MEs and MaEs, 
we calculate their respective recall rates exclusively, without considering the F1-scores.

\subsubsection{Architecture of PESFormer}
In Section \ref{Section 3.2}, our model architecture is primarily constructed by integrating a specific number of component layers: $p_1$ embedding blocks, $p_2$ transformer networks, and $p_3$ 
DTransformer networks. To determine the impact of varying architectural configurations, we establish several combinations in Table \ref{tab1}.

The results presented in Table \ref{tab1} reveal that a moderate increase of the number of embedding blocks (i.e. $p_1$) significantly enhances the model's recall, whereas an excessive $p_1$ leads to 
a decrease in recall. This can be attributed to the fact that too many embedding blocks may introduce noise into the learning process, whereas too few of them may prevent the model from 
capturing fine-grained features. Furthermore, a moderate increase of the number of transformer networks (i.e. $p_2$) notably improves both the recall and precision of our model. However, an excessive 
$p_2$ results in a decrease in both recall and precision. The potential reason for this may be similar to the impact of embedding blocks on the model. Additionally, we observe that adding too 
many DTransformer networks (i.e. $p_3$) significantly degrades the performance of our model. Specifically, the overall precision decreases by 14.6\%, ultimately leading to a reduction in the overall 
F1-score by 11.2\%. Moreover, we find that the combination $[2,2,1]$ yields the best overall precision, overall F1-score, ME recall, and MaE recall. Although the remaining overall recall is not 
the most optimal, it is only 0.6\% worse than the combination $[2,1,1]$ with the best recall.

As for the MaE recall and ME recall, the change follows a similar trend to the overall F1-score. Optimal results can be achieved by moderately adding embedding blocks and transformer 
networks, along with the use of a few DTransformer networks.

\begin{table}[htbp]
  \centering
  \small 
  \setlength\tabcolsep{6pt}
  \caption{Performances with different architectures of PESFormer on the CAS(ME)$^2$ dataset. ``ME REC'' represents the recall rate for MEs, while ``MaE REC'' stands for the recall rate for MaEs. 
          Additionally, ``REC'' denotes the overall recall, ``PRE'' signifies the overall precision rate, and ``F1'' represents the overall F1-score.}
    \begin{tabular}{c|ccccc}
    \hline
    $[p_1,p_2,p_3]$ &ME REC &MaE REC &REC &PRE &F1 \\
    \hline
    $[1,1,1]$       &0.526 &0.770 &0.675 &0.920 &0.779  \\
    $[2,1,1]$       &0.632 &0.827 &\textbf{0.756} &0.918 &0.829  \\                                                                                                                                                                                                                                                                                                                                                                                                                                                                                                                                                                                                                                                                                                                                                                                                                                                                                                                                                                                                                                                                                                                                                                                                                                                                                                                                                                                                                                                                                                                                                                                                                                                                                                                                                                                                                                                                                     
    $[3,1,1]$       &0.579 &0.823 &0.731 &0.919 &0.814  \\
    $[2,2,1]$       &\textbf{0.649} &\textbf{0.840} &0.751 &\textbf{0.947} &\textbf{0.838}  \\
    $[3,2,1]$       &0.632 &0.830 &0.754 &0.937 &0.835  \\
    $[2,2,2]$       &0.404 &0.750 &0.605 &0.908 &0.726  \\
    $[3,2,2]$       &0.526 &0.773 &0.655 &0.880 &0.751  \\
    $[3,3,2]$       &0.421 &0.773 &0.639 &0.844 &0.727  \\
    \hline
    \end{tabular}
  \label{tab1}
\end{table}

\subsubsection{Local Fixed length of Subspaces}
In Section \ref{Section 3.2.3}, DC-MHSA and DCD-MHSA serve to further slice a feature subspace into $U$ temporal subspaces along the temporal dimension. Given that we employ a larger fixed duration 
$G$ as the input video duration for preprocessing, the number of temporal subspaces $U$ is automatically calculated as $U=G/H$ once we specify the local fixed length $H$ within DC-MHSA or DCD-MHSA.

The results in Table \ref{tab2} demonstrate that both excessively large and small values of the $H$ negatively impact the overall recall and precision of our model. Specifically, 
the recall decreases by 5.0\%, while the precision drops by 3.0\% to 5.0\%. Consequently, this also leads to a reduction in the overall F1-score by 4.0\% to 5.0\%. In Section \ref{Section 3.2.1}, 
once we specify the overlapping frames $\varepsilon$ and the number of frames per snippet $s$, the local fixed length $H$ determines the number of frames involved in capturing local longer dependencies 
within a larger temporal subspace. This number of involved frames is calculated as $(H-1)\times(s-\varepsilon)+s$.

In Section \ref{Section 4.3}, we establish the overlapping frames $\varepsilon$ as 6 and set the number of frames per snippet $s$ to 8. When $H$ is set to 19, it indicates that 46 frames (approximately 
1.5 seconds, for a frame rate of 30 FPS) are encompassed for computing local longer dependencies. Based on previous studies such as LSSNet \cite{yu2021lssnet} and LGSNet \cite{yu2024Lgsnet}, 
most facial expressions last for less than 1.5 seconds. Therefore, an excessively large $H$ value tends to bias our model towards favoring proposals with longer durations, while a small 
$H$ biases it towards shorter durations. Furthermore, we find that our model achieves optimal performance with $H=19$ is set to 19.

Regarding the recall of MaEs and MEs, the pattern of change mirrors the trend observed in the overall F1-score. When we set $H$ to 19, the recall for MEs typically outperforms other values of $H$ 
by more than 5.0\%, and similarly, the recall for MaEs generally surpasses other $H$ counterparts by over 10.0\%. This underscores the benefit of selecting a moderate value for $H$ in terms of 
enhancing the computational efficiency of the MHSA blocks and effectively integrating prior information from the distributions within the transformer networks.

\begin{table}[htbp]
  \centering
  \small 
  \setlength\tabcolsep{6pt}
  \caption{Performances with different local fixed lengths of DC-MHSA and DCD-MHSA on the CAS(ME)$^2$ dataset.}
    \begin{tabular}{c|cccccc}
    \hline
    $H$     &ME REC &MaE REC &REC &PRE &F1 \\
    \hline
    9       &0.526 &0.790 &0.700 &0.916 &0.794  \\
    13      &0.509 &0.797 &0.703 &0.909 &0.793  \\                                                                                                                                                                                                                                                                                                                                                                                                                                                                                                                                                                                                                                                                                                                                                                                                                                                                                                                                                                                                                                                                                                                                                                                                                                                                                                                                                                                                                                                                                                                                                                                                                                                                                                                                                                                                                                                                                     
    19      &\textbf{0.649} &\textbf{0.840} &\textbf{0.751} &\textbf{0.947} &\textbf{0.838}  \\
    25      &0.561 &0.780 &0.695 &0.929 &0.795  \\
    37      &0.526 &0.783 &0.703 &0.900 &0.789  \\
    \hline
    \end{tabular}
  \label{tab2}
\end{table}

\begin{table*}[htbp]
  \centering
  \small 
  \setlength\tabcolsep{13pt}
  \caption{Performances with different encoding methods on the CAS(ME)$^2$ dataset.}
    \begin{tabular}{c|c|c|ccccc}
    \hline
    Model &Preprocess &Encoding Method  &ME REC&MaE REC &REC &PRE &F1 \\
    \hline
    \multirow{4}[2]{*}{CNN}     &Sliding Window  &Anchor &0.211  &0.487  &0.398  &0.481  &0.436  \\
                                &Sliding Window  &DTE    &0.404  &0.707  &0.524  &0.728  &0.609  \\
                                &Fixed Duration  &Anchor &0.053  &0.177  &0.078  &0.129  &0.098  \\
                                &Fixed Duration  &DTE    &0.281  &0.540  &0.398  &0.740  &0.517  \\
    \hline
    \multirow{4}[2]{*}{Trans}   &Sliding Window  &Anchor &0.193  &0.557  &0.417  &0.482  &0.447  \\
                                &Sliding Window  &DTE    &0.561  &0.797  &0.658  &0.874  &0.751  \\
                                &Fixed Duration  &Anchor &0.105  &0.350  &0.305  &0.319  &0.312  \\
                                &Fixed Duration  &DTE    &\textbf{0.649} &\textbf{0.840} &\textbf{0.751} &\textbf{0.947} &\textbf{0.838}  \\
    \hline
    \end{tabular}
  \label{tab3}
\end{table*}

\subsubsection{Encoding Method and Preprocessing}
\label{Section 4.4.3}
In Section \ref{Section 3.3}, we emphasize the advantages of our proposed DTE encoding method while discussing the potential problems associated with the anchor encoding method. To assess the 
effectiveness of our methods, we utilize both convolutional neural networks (CNN)-based \cite{yu2024Lgsnet} and transformer-based models. Additionally, we conduct an ablation analysis using the 
two preprocessing techniques described in Section \ref{Section 3.5}. 

Regarding the encoding method, the results in Table \ref{tab3} clearly demonstrate that, given the same model and preprocessing technique, our DTE method significantly outperforms the anchor 
method in terms of overall recall, overall precision, and F1-score. Specifically, the overall recall improves by at least 12.0\%, the overall precision by at least 9.0\%, and the F1-score by at 
least 17.0\%. Notably, when using the transformer model with larger fixed duration preprocessing, the improvement in F1-score using our DTE method exceeds 50.0\% compared with the anchor method. 
This validates the effectiveness of our DTE method, particularly when employed with transformer-based models.

Regarding the model performance, our experiments reveal that the transformer-based model outperforms the CNN-based model when using the same encoding and preprocessing methods. This finding 
suggests that transformer-based models are more effective at capturing the temporal dependencies inherent in sequential data, leading to improved performance.

Regarding the preprocessing, our investigation reveals that the CNN-based model fails to yield significant improvements in overall recall, overall precision, and F1-score when utilizing the larger 
fixed duration method as opposed to sliding windows during preprocessing, given a constant encoding method. This observation can be attributed to the fact that sliding windows during preprocessing 
for CNN-based models increases the number of training intervals, enabling the model to learn more representative features. Conversely, the larger fixed duration method in the CNN-based models may 
restrict the model's ability to capture the complete contextual information present in the input data, thereby limiting its performance. In contrast the combination of the DTE method and the larger fixed 
duration preprocessing achieves optimal results when used with transformer-based models. This is because the larger fixed duration preprocessing ensures a longer training video 
duration, which requires our well-designed transformer-based model to extract longer dependencies and fine-grained information. This extracted information is crucial for ensuring that the DTE encoding 
method achieves its best performance.

Moreover, our DTE encoding method significantly enhances both MaE recall and ME recall by at least 21.0\% and 19.0\%, respectively. This further validates the effectiveness of the DTE method and 
transformer-based model in expression spotting tasks. Additionally, the preprocessing technique also exhibits similar trends in improving the F1-score, which further demonstrates the importance of 
preprocessing techniques in improving model performance.

\subsubsection{Different Portfolios of Losses}
In Section \ref{Section 3.4}, we employ both the focal loss function \cite{lin2017focal} and the dice loss function \cite{milletari2016v} to refine our model. The results in Table \ref{tab4} 
demonstrate that incorporating the dice loss function significantly boosts the overall recall by 2.9\%, precision by 2.6\%, and F1-score by 2.8\%. The primary reason for this improvement lies 
in the imbalance distribution of the datasets. Although the focal loss is effective in mitigating sample imbalance, the distribution of foreground content within the videos remains highly skewed, 
with some videos exhibiting foreground shares below 1.0\%. The dice loss, on the other hand, is specifically designed to tackle extreme sample imbalance in medical image segmentation tasks. 
Consequently, the integration of dice loss into our model is particularly beneficial in addressing this imbalance, particularly in scenarios characterized by sparse foreground content.

With regard to MaE recall and ME recall, the inclusion of dice loss results in significant improvements, boosting them by 4.0\% and 7.0\%, respectively. This indicates that incorporating dice loss 
enhances the accuracy of proposals captured by the model. Notably, the improvement is more pronounced in shorter proposal intervals, suggesting that dice loss is particularly effective in capturing 
more accurate proposals. Additionally, the dice loss effectively mitigates the imbalance between the foreground and background, further optimizing the model's performance.

\begin{table}[htbp]
  \centering
  \small 
  \setlength\tabcolsep{6pt}
  \caption{Performances with different portfolios of losses on the CAS(ME)$^2$ dataset.}
    \begin{tabular}{c|cccccc}
    \hline
    Loss        &ME REC&MaE REC &REC &PRE &F1 \\
    \hline
    Focal       &0.579 &0.800 &0.723 &0.921 &0.810  \\
    Focal+Dice  &\textbf{0.649} &\textbf{0.840} &\textbf{0.751} &\textbf{0.947} &\textbf{0.838}  \\                                                                                                                                                                                                                                                                                                                                                                                                                                                                                                                                                                                                                                                                                                                                                                                                                                                                                                                                                                                                                                                                                                                                                                                                                                                                                                                                                                                                                                                                                                                                                                                                                                                                                                                                                                                                                                                                                    
    \hline
    \end{tabular}
  \label{tab4}
\end{table}

\subsubsection{Effect of Different Loss Weight}
In Equation \ref{equ3.10}, we introduce the loss weight $\lambda$ as a means to balance the various components of our loss function. As evident from Table \ref{tab5}, our model exhibits varying 
performance across a wide range of $\lambda$ values. Specifically, we observe a maximum gap of 7.6\% in overall recall, 5.8\% in overall precision, 6.2\% in F1-score, 10.5\% in ME recall and 4.3\% 
in MaE recall. Upon careful evaluation, we find that setting $\lambda$ to 1.0 yields the optimal results. 
Therefore, in all our experiments, we adopt $\lambda=1.0$ as the default value to ensure consistent and reliable performance.

\begin{table}[htbp]
  \centering
  \small 
  \setlength\tabcolsep{6pt}
  \caption{Performances with different loss weights on the CAS(ME)$^2$ dataset.}
    \begin{tabular}{c|cccccc}
    \hline
    $\lambda$ &ME REC &MaE REC &REC &PRE &F1 \\
    \hline
    0.25      &0.544 &0.753 &0.675 &0.913 &0.776  \\
    0.5       &0.526 &0.770 &0.695 &0.902 &0.785  \\                                                                                                                                                                                                                                                                                                                                                                                                                                                                                                                                                                                                                                                                                                                                                                                                                                                                                                                                                                                                                                                                                                                                                                                                                                                                                                                                                                                                                                                                                                                                                                                                                                                                                                                                                                                                                                                                                     
    1.0       &\textbf{0.649} &\textbf{0.840} &\textbf{0.751} &\textbf{0.947} &\textbf{0.838}  \\
    2.0       &0.509 &0.797 &0.720 &0.889 &0.796  \\
    \hline
    \end{tabular}
  \label{tab5}
\end{table}

\subsubsection{Effect of Different Fixed Duration}
In Section \ref{Section 4.4.3}, we have demonstrated that using a larger fixed duration outperforms the sliding windows. Additionally, in Section \ref{Section 3.5}, we 
establish a larger fixed duration $G$ for preprocessing the input video features to standardize the duration of training videos. To further investigate the impact of different $G$ values on model 
performance, we conducted experiments with a range of $G$ values: $[144, 288, 576, 1152, 2304, 3456]$ as presented in Table \ref{tab6}.

The results indicate that as the fixed duration $G$ increases, the overall recall and overall precision of the model generally improve, with a corresponding increase in the F1-score. However, a notable 
turning point is observed at $G=2304$, where recall, precision, and F1-score peak at 75.1\%, 94.7\%, and 83.8\% respectively. Beyond this point, further increasing $G$ leads to a decline in all three 
metrics. Additionally, we observe significant variations in performance, with a maximum gap of 12.3\% in ME recall and 4.0\% in MaE recall across different $G$ values.

The primary reason for the above results is that a larger zero-complement area hinders the model's capacity to learn fine-grained features and introduces substantial noise, ultimately compromising its 
performance. Therefore, in all our experiments, we adopt $G=2304$ as the default value. 

\begin{table}[htbp]
  \centering
  \small 
  \setlength\tabcolsep{6pt}
  \caption{Performances with different fixed durations on the CAS(ME)$^2$ dataset.}
    \begin{tabular}{c|cccccc}
    \hline
    $G$ &ME REC &MaE REC &REC &PRE &F1 \\
    \hline
    144       &0.614 &0.793 &0.641 &0.779 &0.704  \\
    288       &0.526 &0.773 &0.650 &0.862 &0.741  \\                                                                                                                                                                                                                                                                                                                                                                                                                                                                                                                                                                                                                                                                                                                                                                                                                                                                                                                                                                                                                                                                                                                                                                                                                                                                                                                                                                                                                                                                                                                                                                                                                                                                                                                                                                                                                                                                                     
    576       &0.596 &0.793 &0.686 &0.904 &0.780  \\
    1152      &0.544 &0.797 &0.714 &0.895 &0.794  \\
    2304      &\textbf{0.649} &\textbf{0.840} &\textbf{0.751} &\textbf{0.947} &\textbf{0.838}  \\
    3456      &0.561 &0.800 &0.720 &0.902 &0.801  \\
    \hline
    \end{tabular}
  \label{tab6}
\end{table}

\subsubsection{Effect of Different Truncation Threshold}
In Section \ref{Section 3.6}, we establish a truncation threshold $\theta$ to select timestamp snippets that are larger than $\theta$ as valid snippets for generating proposal. To assess the impact of varying 
thresholds on model performance, we tested $\theta$ ranging from 0.005 to 0.1. The results presented in Table \ref{tab7} indicate that a higher $\theta$ generally leads to improved overall recall, overall 
precision, and F1-score. Notably, when setting the threshold at or above 0.05, our model exhibits minimal performance gaps, with only 1.7\% in overall recall, 0.1\% in overall precision, and 1.2\% in F1-score. 
Therefore, to maintain consistent and optimal performance, we adopt $\theta=0.05$ as the default value in all our experiments.

\begin{table}[htbp]
  \centering
  \small 
  \setlength\tabcolsep{6pt}
  \caption{Performances with different truncation threshold on the CAS(ME)$^2$ dataset.}
    \begin{tabular}{c|cccccc}
    \hline
    $\theta$ &ME REC &MaE REC &REC &PRE &F1 \\
    \hline
    0.005     &0.526 &0.790 &0.703 &0.930 &0.801  \\
    0.01      &0.596 &0.823 &0.711 &0.901 &0.795  \\                                                                                                                                                                                                                                                                                                                                                                                                                                                                                                                                                                                                                                                                                                                                                                                                                                                                                                                                                                                                                                                                                                                                                                                                                                                                                                                                                                                                                                                                                                                                                                                                                                                                                                                                                                                                                                                                                     
    0.02      &0.579 &0.813 &0.723 &0.908 &0.805 \\
    0.05      &\textbf{0.649} &\textbf{0.840} &\textbf{0.751} &\textbf{0.947} &\textbf{0.838}  \\
    0.1       &0.614 &0.817 &0.734 &0.946 &0.826  \\
    \hline
    \end{tabular}
  \label{tab7}
\end{table}

\begin{table}[htbp]
  \centering
  \small 
  \setlength\tabcolsep{2pt}
  \caption{Comparison with the state-of-the-art models on the CAS(ME)$^2$ dataset.}
    \begin{tabular}{c|cccccc}
    \hline
    Model &ME REC &MaE REC &REC &PRE &F1 \\
    \hline
    He et al. (2020)\cite{he2020spotting}     &   -  &  -   &0.020 &0.364 &0.038 \\
    Zhang et al. (2020)\cite{zhang2020spatio} &   -  &  -   &0.085 &0.406 &0.140 \\
    MESNet (2021) \cite{wang2021mesnet}       &   -  &  -   &  -   &   -  &0.036 \\
    Yap et al. (2021) \cite{yap20213dcnn}     &   -  &  -   &  -   &   -  &0.030 \\
    LSSNet (2021) \cite{yu2021lssnet}         &   -  &  -   &  -   &   -  &0.327 \\
    He et al. (2021) \cite{yuhong2021research}&   -  &  -   &  -   &   -  &0.343 \\
    MTSN (2022) \cite{liong2022mtsn}          &   -  &  -   &0.342 &0.385 &0.362 \\
    Zhao et al. (2022) \cite{zhao2022rethink} &   -  &   -  &   -  &   -  &0.403 \\
    LGSNet (2024)\cite{yu2024Lgsnet}          &0.221 &0.513 &0.367 &0.630 &0.464 \\
    PESFormer                                &\textbf{0.649} &\textbf{0.840} &\textbf{0.751} &\textbf{0.947} &\textbf{0.838} \\
    \hline
    \end{tabular}
  \label{tab8}
\end{table}

\begin{table}[htbp]
  \centering
  \small 
  \setlength\tabcolsep{2pt}
  \caption{Comparison with the state-of-the-art models on the SAMM-LV dataset.}
    \begin{tabular}{c|cccccc}
    \hline
    Model &ME REC &MaE REC &REC &PRE &F1 \\
    \hline
    He et al.  (2020) \cite{he2020spotting}   &   -  &   -  &0.029 &0.101 &0.045 \\
    Zhang et al. (2020) \cite{zhang2020spatio}&   -  &   -  &0.079 &0.136 &0.100 \\
    MESNet (2021) \cite{wang2021mesnet}       &   -  &   -  &   -  &  -   &0.088 \\
    Yap et al. (2021) \cite{yap20213dcnn}     &   -  &   -  &   -  &  -   &0.119 \\
    LSSNet (2021) \cite{yu2021lssnet}         &   -  &   -  &   -  &  -   &0.290 \\
    He et al. (2021) \cite{yuhong2021research}&   -  &   -  &   -  &  -   &0.364 \\
    MTSN (2022) \cite{liong2022mtsn}          &   -  &   -  &0.260 &0.319 &0.287 \\
    Zhao et al. (2022) \cite{zhao2022rethink} &   -  &   -  &   -  &   -  &0.386 \\
    LGSNet (2024)\cite{yu2024Lgsnet}          &0.257 &0.487 &0.355 &0.429 &0.388 \\
    PESFormer                                &\textbf{0.566} &\textbf{0.585} &\textbf{0.523} &\textbf{0.864} &\textbf{0.652} \\
    \hline
    \end{tabular}
  \label{tab9}
\end{table}

\begin{table}[htbp]
  \centering
  \small 
  \setlength\tabcolsep{2pt}
  \caption{Comparison with the state-of-the-art models on the CAS(ME)$^3$ dataset.}
    \begin{tabular}{c|cccccc}
    \hline
    Model &ME REC &MaE REC &REC &PRE &F1 \\
    \hline
    LGSNet (2024)\cite{yu2024Lgsnet}      &0.166 &0.416 &0.292 &0.196 &0.235 \\
    PESFormer                            &\textbf{0.643} &\textbf{0.803} &\textbf{0.746} &\textbf{0.960} &\textbf{0.839} \\
    \hline
    \end{tabular}
  \label{tab10}
\end{table}

\subsection{Comparison with State-of-the-art Methods}
We comprehensively compared our PESFormer with recent state-of-the-art methods \cite{he2020spotting, zhang2020spatio, yu2024Lgsnet} across three datasets: CAS(ME)$^2$, CAS(ME)$^3$ and SAMM-LV. The results, 
summarized in Tables \ref{tab8}, \ref{tab9}, and \ref{tab10}, demonstrate the clear superiority of PESFormer over existing spotting methods.

On the CAS(ME)$^2$ dataset, PESFormer outperforms previous approaches by at least 48.8\% in overall recall, 31.7\% in overall precision, 37.4\% in F1-score, 32.7\% in MaE recall, and 44.8\% in ME recall.
Similarly, on the SAMM-LV dataset, PESFormer exhibits significant advantages, achieving improvements of at least 16.8\% in overall recall, 43.5\% in overall precision, 26.4\% in F1-score, 9.8\% in MaE 
recall, and 30.9\% in ME recall. Lastly, on the CAS(ME)$^3$ dataset, PESFormer demonstrates even more pronounced superiority, with improvements of at least 45.4\% in overall recall, 76.4\% in overall 
precision, 60.4\% in F1-score, 38.8\% in MaE recall, and 47.7\% in ME recall.

These results consistently highlight the effectiveness of PESFormer in comparison to existing methods for MaE and ME spotting. However, we find that when training PESFormer on the SAMM-LV dataset, the 
improvement is still relatively than when PESFormer is trained on the CAS(ME)$^2$ and CAS(ME)$^3$ datasets. The main reason is that some longer ground truth intervals of the SAMM-LV dataset last longer than 4.0 
seconds, and some long ones even exceed 10.0 seconds. This distribution is clearly inconsistent with Ekman's observation that the duration of MEs is less than 0.5 second and that the duration of MaEs ranges 
between 0.5 and 4 seconds \cite{ekman2003darwin}. In addition, we also find that when training PESFormer on the CAS(ME)$^3$ dataset, the improvement is more significant than when PESFormer is trained on 
the CAS(ME)$^2$ and SAMM-LV datasets. The main reason is that although the CAS(ME)$^3$ dataset contains more training videos, a wider data distribution, and more variable expressions. 
In total, the above results show that our PESFormer captures more effective features as well as achieves better spotting results compared to existing methods.

\section{Conclusion}
In this paper, we introduce a novel model called PESFormer, grounded on the ViT architecture. PESFormer comprises the fundamental components of embedding blocks, transformer blocks, and output blocks, 
adopting a streamlined approach that necessitates only two simple loss functions and a single classification output. Furthermore, we employ the strategy of direct timestamp encoding (DTE) for each snippet
without focusing on 
specific onset and offset timestamps of ground truth expressions. This approach significantly mitigates the challenges associated with model learning and proposal generation. For preprocessing simplicity,
we set a larger fixed duration for untrimmed videos, complementing it with zero padding, rather than the conventional method of splitting original videos into short-duration sliding windows. We conducted 
extensive experiments on the CAS(ME)$^2$, CAS(ME)$^3$, and SAMM-LV datasets to comprehensively demonstrate the effectiveness of our model. 

However, one limitation of our method lies in the need for further improvements on the SAMM-LV dataset. A potential approach to address this limitation is to develop an efficient module capable of capturing 
more granular information to accommodate more widely distributed data.

\section*{Acknowledgments}
This study was supported by the STI2030-Major Projects (2022ZD0204600) and the National Natural Science Foundation of China (62476050).  
This work was also partly supported by Huzhou Science and Technology Program (\#2023GZ13).

\bibliographystyle{IEEEtran}
\bibliography{ref}

\begin{IEEEbiography}[{\includegraphics[width=1in,height=1.25in,clip,keepaspectratio]{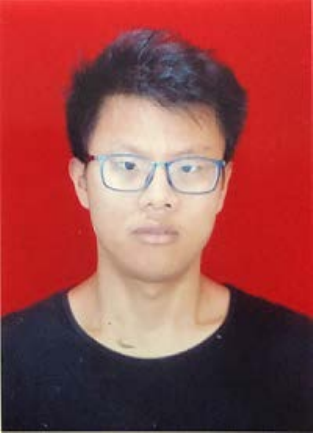}}]
  {Wang-Wang Yu} received the M.S. degree in biomedical engineering from University of Electronic Science and Technology of China (UESTC) in 2020. 
  He is now pursuing his Ph.D. degree in UESTC. His research interests include video understanding, emotional analysis, weakly supervised learning.
\end{IEEEbiography}

\begin{IEEEbiography}[{\includegraphics[width=1in,height=1.25in,clip,keepaspectratio]{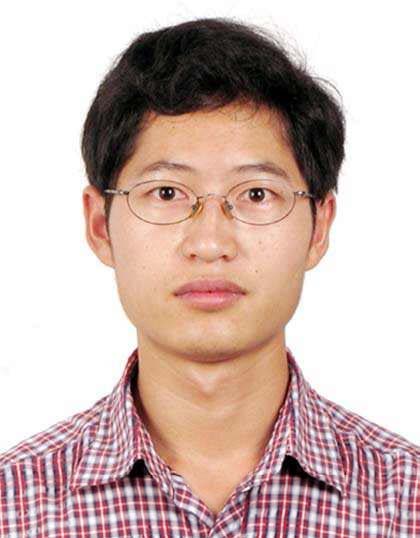}}]
  {Kai-Fu Yang} received the Ph.D. degree in biomedical engineering from the University of Electronic Science and 
  Technology of China (UESTC), Chengdu, China, in 2016. He is currently an associate research professor with 
  the MOE Key Lab for Neuroinformation, School of Life Science and Technology, UESTC, Chengdu, China. His research 
  interests include cognitive computing and brain-inspired computer vision.
\end{IEEEbiography}

\begin{IEEEbiography}[{\includegraphics[width=1in,height=1.25in,clip,keepaspectratio]{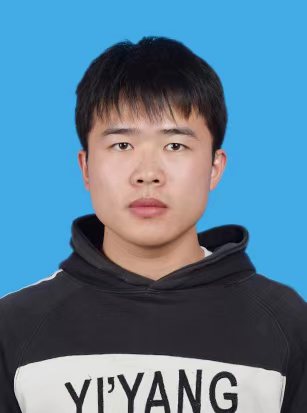}}]
  {Xiangrui Hu} earned his Bachelor's degree from Dalian Ocean University and is currently pursuing his Master's degree at the University of Electronic 
  Science and Technology of China (UESTC). His research interests include object detection and image enhancement.
\end{IEEEbiography}

\begin{IEEEbiography}[{\includegraphics[width=1in,height=1.25in,clip,keepaspectratio]{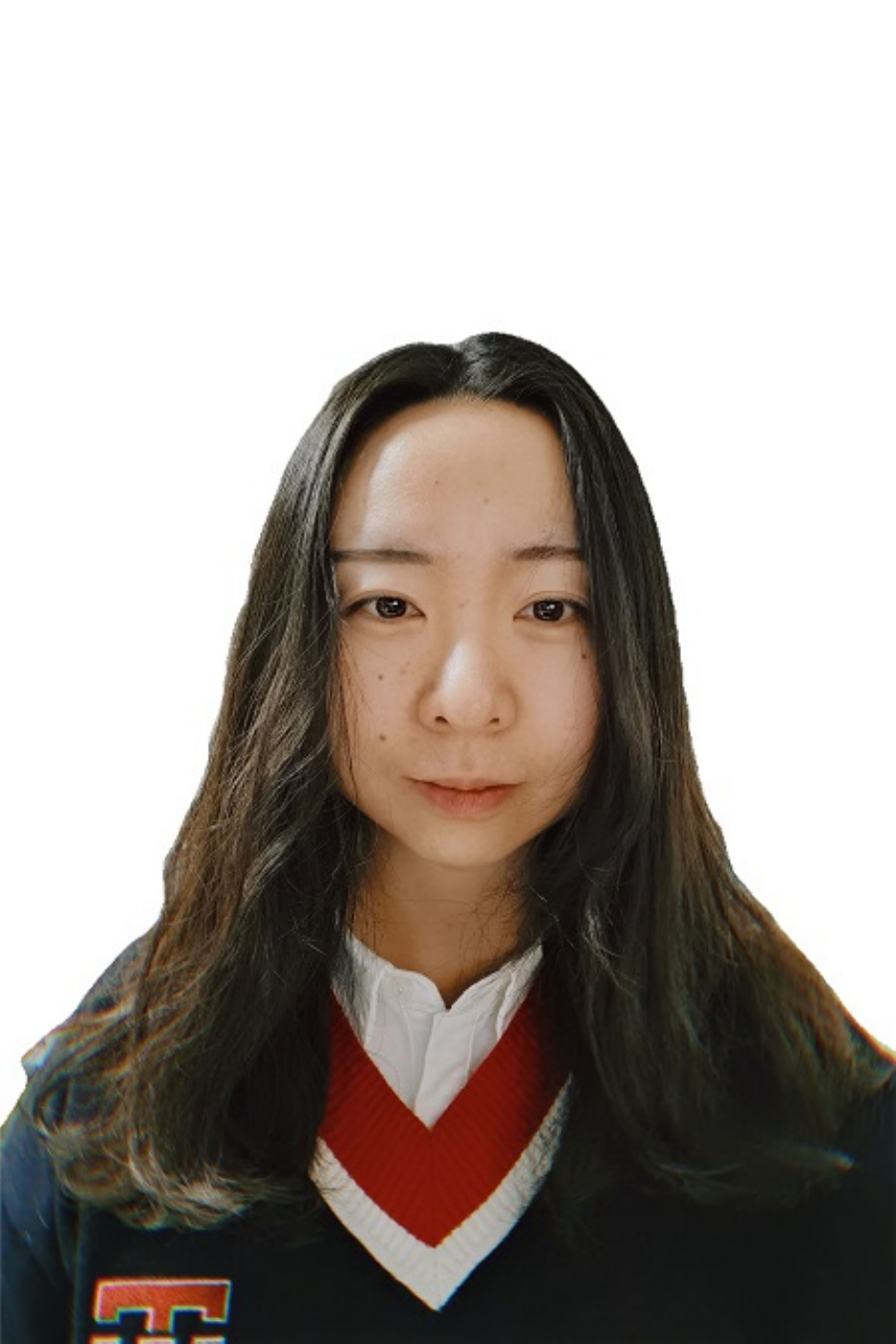}}]
  {Jingwen Jiang} received the M.S. degree in Statistics from Soochow University, in 2015. She is now
  pursuing her Ph.D. degree in Sichuan University. She research interests include emotion recognition.
\end{IEEEbiography}

\begin{IEEEbiography}[{\includegraphics[width=1in,height=1.25in,clip,keepaspectratio]{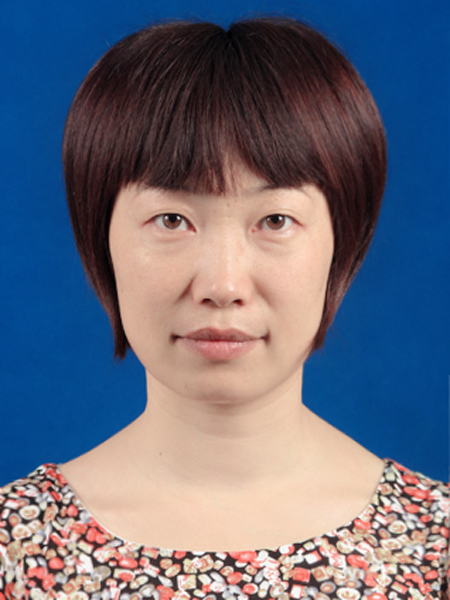}}]
  {Hong-Mei Yan} received the Ph.D. degree from Chongqing University in 2003. She is now a Professor 
  with the MOE Key Laboratory for Neuroinformation, University of Electronic Science and Technology of China, Chengdu, China. 
  Her research interests include visual cognition, visual attention, visual encoding and decoding.
\end{IEEEbiography}

\begin{IEEEbiography}[{\includegraphics[width=1in,height=1.25in,clip,keepaspectratio]{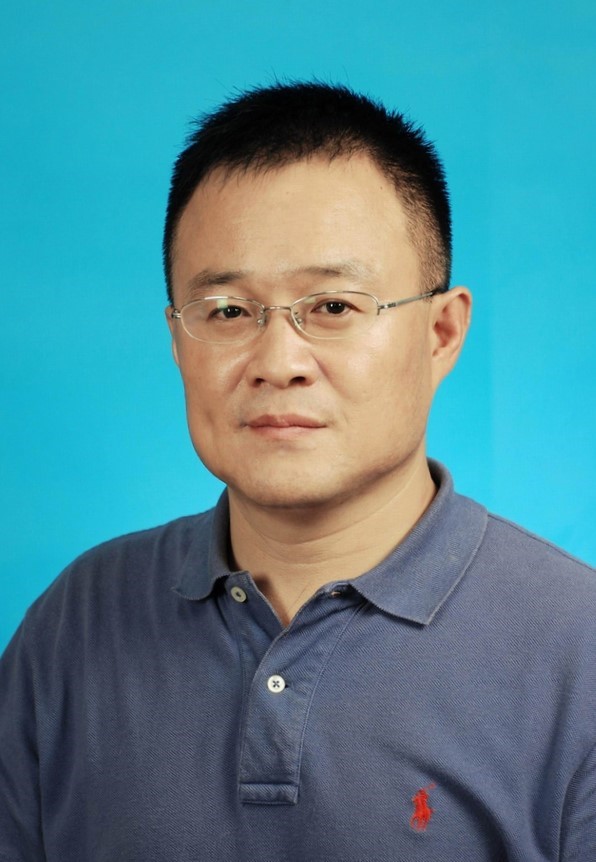}}]
  {Yong-Jie Li} (Senior Member, IEEE) received the Ph.D. degree in biomedical engineering from
  UESTC, in 2004. He is currently a Professor with the Key Laboratory for NeuroInformation of Ministry of
  Education, School of Life Science and Technology, University of Electronic Science and Technology of
  China. His research focuses on building of biologically inspired computational models of visual perception and 
  the applications in image processing and computer vision.
\end{IEEEbiography}

\end{document}